\pdfoutput=1

\documentclass[letterpaper, 10 pt, conference]{ieeeconf}  

\IEEEoverridecommandlockouts                              

\usepackage{color}

\usepackage{amsmath} 
\usepackage{graphicx}
\usepackage{capt-of}   
\usepackage{cuted} 

\usepackage[hidelinks]{hyperref}

\usepackage{tikz}
\usepackage{pgfplots}
\usepackage{pifont}
\newcommand{\cmark}{\ding{51}} 
\newcommand{\xmark}{\ding{55}} 
\pgfplotsset{compat=1.18}
\usepackage{algorithm}
\usepackage{algpseudocode} 
\floatname{algorithm}{Protocol}

\AtBeginDocument{%
  \setcounter{dbltopnumber}{5}

  \setlength{\dbltextfloatsep}{8pt plus 2pt minus 2pt}
  \setlength{\dblfloatsep}{8pt plus 2pt minus 2pt}
}

\title{\LARGE \bf
A Multimodal Data Collection Framework for Dialogue-Driven Assistive Robotics to Clarify Ambiguities: A Wizard-of-Oz Pilot Study}

\author{Guangping Liu$^{1}$, Nicholas Hawkins$^{1}$, Billy Madden$^{1}$, Tipu Sultan$^{1}$, Flavio Esposito$^{2}$, Madi Babaiasl$^{1}$ 
\thanks{$^{1}$Aerospace and Mechanical Engineering Department, $^{2}$Computer Science Department, Saint Louis University, St. Louis, MO, 63103, United States. Dr. Madi Babaiasl is the corresponding author of this paper ({\tt\small madi.babaiasl@slu.edu}).}%
\thanks{*This work has been supported by SLU's Research Institute under Award-01935, Clare Boothe Luce Foundation under Award-01281, and SLU's AEME department under Proj-000480.}%
}

\begin{document}
\maketitle

\begin{strip}
\vspace{-24mm} 
\centering
\includegraphics[width=\textwidth]{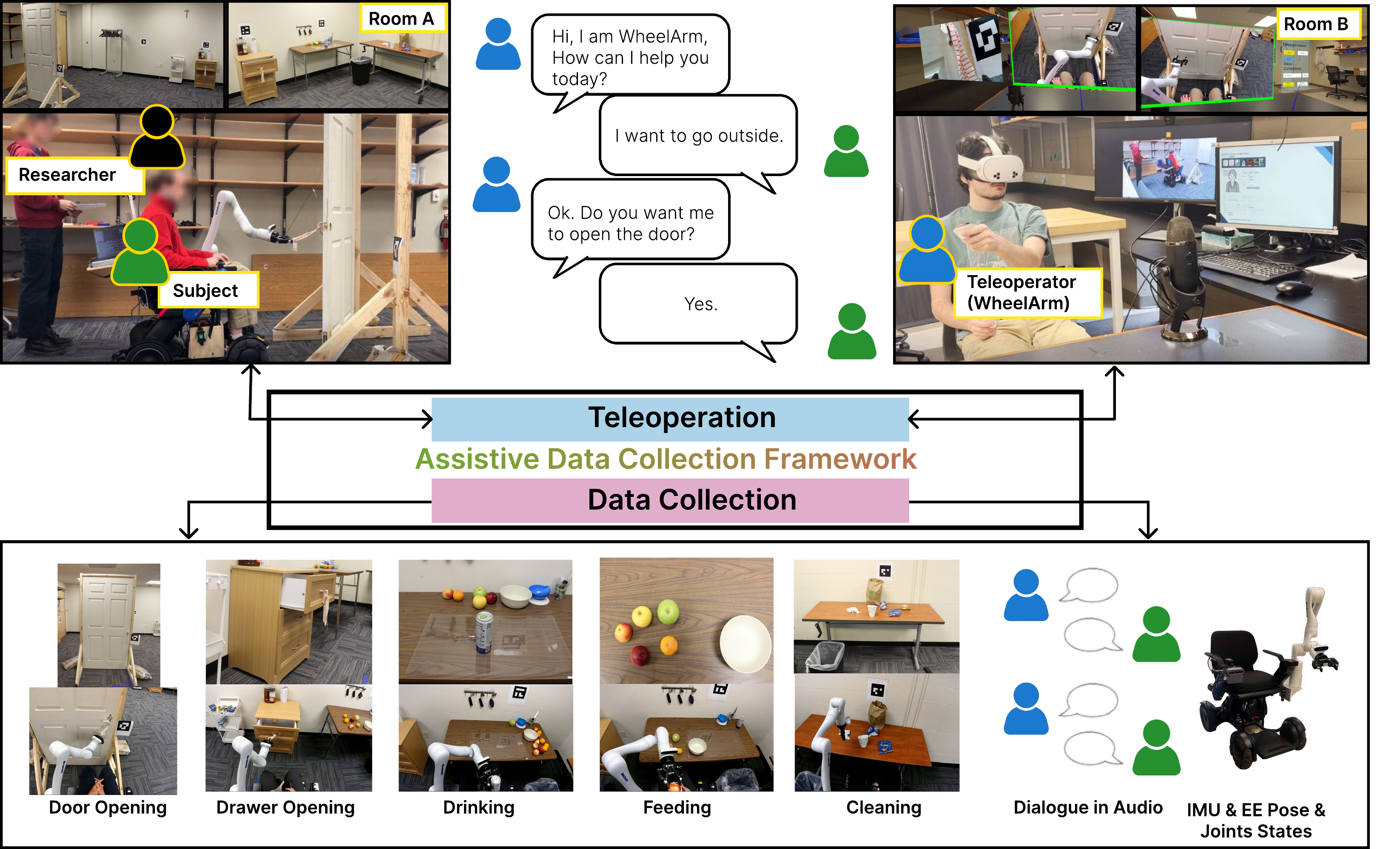}
\vspace{-1.5em}  
\captionof{figure}{\textbf{Overview of the Multimodal Assistive Data Collection Framework:}
Our assistive data collection framework comprises a Virtual Reality (VR) based teleoperation system for a wheelchair and robotic arm, and a real-time multimodal data recording pipeline. The experimental setup spans two physically separated spaces (Room~A and Room~B). Room~A is arranged with five assistive tasks, including door opening, drawer opening, drinking, feeding, and cleaning. Room~B houses the teleoperator, who follows a Wizard-of-Oz (WoZ) protocol \cite{riek2012wizard} to simulate robot autonomy and elicit natural dialogue. The resulting dataset contains 53 trials with five synchronized modalities, including RGB-D video, human-robot conversational audio, Inertial Measurement Unit (IMU) measurements, and robot kinematics, which are end-effector (EE) poses and whole-body joint states.}
\label{fig:overview}
\vspace{-4mm}
\end{strip}

\section{Abstract}
\label{sec:abstract}

\begin{table*}[!t]
\centering
\caption{Comparison of Existing Multimodal Frameworks \& Datasets in Assistive Robotics}
\label{tab:comparison}
\begin{tabular}{llllccll}
\hline
Data Collection Frameworks and Datasets & Emb.\textsuperscript{1} & Interact.\textsuperscript{2} & TT.\textsuperscript{3} & Ambig.\textsuperscript{4} & WoZ & Modal.\textsuperscript{5} & Dom.\textsuperscript{6}\\
\hline
RT-1~\cite{brohan2022rt} & ER & SI & M, N & \xmark & \xmark & RGB, T, C & R\\
BridgeData V2~\cite{walke2023bridgedata} & W & SI & M & \xmark & \xmark & RGB-D, T, C & R\\
ALFRED~\cite{shridhar2020alfred} & NE & MD & M, N &\xmark & \xmark &I, T, C & S\\
DialFRED~\cite{gao2022dialfred} & NE & MD & M, N & \cmark & \xmark & I, T, C & S\\
DROID~\cite{khazatsky2024droid} & F & MD & M & \xmark & \xmark & RGB-D, T, C, J & R\\
\hline
Ours & WH, K & MD & M, N &\cmark & \cmark & RGB-D, A, T, C, J & R\\
\hline
\end{tabular}
\vspace{1mm}
\begin{minipage}{\textwidth}
\footnotesize
\textsuperscript{1}Embodiment. (ER: Everyday Robots. W:  WidowX 250. NE: No Embodiment. F: Franka. WH: Whill Model CR2. K: Kinova Gen3 Robotic Arm)
\textsuperscript{2}Interaction Type. (SI: Single Turn Instruction. MD: Multiturn Dialogues.
\textsuperscript{3}Task Type. (M: Manipulation. N: Navigation)
\textsuperscript{4}Ambiguity Support.
\textsuperscript{5}Modalities Captured. (A: Audio. T: Text. C: Cartesian Pose. J: Joint States. I: Images.
\textsuperscript{6}Data Domain. (S: Simulation Data. R: Real World Data.)
\end{minipage}
\vspace{-0.5em}
\end{table*}

Integrated control of wheelchairs and wheelchair-mounted robotic arms (WMRAs) has strong potential to increase independence for users with severe motor limitations, yet existing interfaces often lack the flexibility needed for intuitive assistive interaction. Although data-driven AI methods show promise, progress is limited by the lack of multimodal datasets that capture natural Human–Robot Interaction (HRI), particularly conversational ambiguity in dialogue-driven control. To address this gap, we propose a multimodal data collection framework that employs a dialogue-based interaction protocol and a two-room Wizard-of-Oz (WoZ) setup to simulate robot autonomy while eliciting natural user behavior. The framework records five synchronized modalities: RGB-D video, conversational audio, inertial measurement unit (IMU) signals, end-effector Cartesian pose, and whole-body joint states across five assistive tasks. Using this framework, we collected a pilot dataset of 53 trials from five participants and validated its quality through motion smoothness analysis and user feedback. The results show that the framework effectively captures diverse ambiguity types and supports natural dialogue-driven interaction, demonstrating its suitability for scaling to a larger dataset for learning, benchmarking, and evaluation of ambiguity-aware assistive control. The dataset and codes will be released at \href{https://madibabaiasl.github.io/WheelArmWoZDataset/}{https://madibabaiasl.github.io/WheelArmWoZDataset/} upon paper acceptance, and a demonstration video is available at \href{https://youtu.be/4Ei7vba7TNY}{https://youtu.be/4Ei7vba7TNY}.

\vspace{-0.5em}

\section{Introduction}
\label{sec:introduction}

Individuals with severe limb limitations, such as those with spinal cord injury (SCI), increasingly rely on advanced assistive technologies like wheelchairs and Wheelchair-mounted Robotic Arms (WMRAs) to regain independence in Activities of Daily Living (ADLs). According to the Centers for Disease Control and Prevention (CDC)'s report, 28.7\% of adults in the United States have disabilities, of which 13.9\% have cognitive issues, 12.2\% have mobility problems, and 7.7\% have difficulties in independent living~\cite{CDCDisabilityInfographic}. These statistics show the need for assistive devices that are intuitive and easy to use. However, current interactions are limited to joysticks~\cite{rulik2022control}, touchscreen~\cite{chung2017performance}, or a restricted set of predefined voice commands~\cite{poirier2023efficient}, which impose a significant cognitive and physical burden on the user~\cite{chung2017performance, chung2023robotic}. These usage difficulties and lack of intuitive control can lead to user frustration and heavy cognitive load, creating a critical gap between robotic capabilities and real-world usability.

Recent advances in artificial intelligence provide a potential solution to replace cumbersome interfaces with natural daily conversations. A dialogue-based interaction powered by large language models (LLMs) can understand vagueness in human daily conversation by asking questions to resolve different types of ambiguities, such as intent or spatial ambiguities~\cite{ramrakhya2025grounding, lin2025ask}.  However, there is a lack of datasets that encompass human-robot conversational interaction, vision data, and robotic data in real-world assistive tasks. This gap heavily constrains the development of intuitive and intelligent assistive robot control.

To bridge this gap, we present a data collection framework using WoZ that captures natural dialogue, video, and robot telemetry during five essential assistive tasks (see \textbf{Fig.~\ref{fig:overview}}). Our framework captures multimodal data, the nuance of natural conversational interaction, and elicits authentic user dialogue. We collected 53 trials and assessed usability through post-study questionnaires on subjects' feedback of the dialogue-driven interactional intelligent assistive robot.

In summary, this work makes the following contributions: 
\begin{itemize} 
\item A high-fidelity pilot multimodal real-world dataset to prove the high quality of our data collection methods: We provide a synchronized dataset capturing natural human and robot interaction dialogues, ego-centric and wrist vision, IMU, end effector cartesian pose, and whole-body joint states for assistive ADLs. 
\item A data collection framework supporting WoZ: We present a novel technical integration that extends OpenTeach~\cite{iyer2024open}, originally designed for arm manipulation, to enable simultaneous control of a wheelchair and a robotic arm. Additionally, the VR application is customized in Unity to include additional control functionality to reduce the risk of teleoperation failure.  
\item Validate an intelligent robot interaction protocol: We establish a conversational interaction protocol for assistive robot control and collect preliminary user feedback demonstrating the enjoyment of natural conversational interaction over traditional joystick interfaces. \end{itemize}

To the best of our knowledge, our work is the first dataset to capture synchronized natural dialogue, wrist and ego-centered cameras, and whole-body robot states for mobile assistive tasks in real-world settings using the WoZ method. Our work will significantly accelerate the intelligent control using artificial intelligence methods on mobile manipulators, wheelchairs, robotic arms, and other similar assistive devices. 
\vspace{-1.2em}



\section{Related Work}
\label{sec:related work}\subsection{Multimodal Datasets for Assistive Robotics}
With the rise of large foundation models, multimodal robot datasets have become central to data-driven learning. Large-scale datasets such as Open X-Embodiment, DROID, and BridgeData V2~\cite{o2024open, khazatsky2024droid, walke2023bridgedata} span diverse robot embodiments but largely focus on manipulation. In the assistive domain, OpenRoboCare~\cite{liang2025openrobocare} and Harmonic~\cite{newman2022harmonic} provide rich human-centered sensing, such as eye tracking, yet typically lack detailed robot data, such as end-effector data and none has realistic human-robot dialogue data (see \textbf{Table \ref{tab:comparison}}).
\vspace{-0.5em}

\subsection{Natural Language and Dialogue in HRI}
Natural-language interaction in assistive HRI has been explored through direct devices, such as touchscreens and joysticks~\cite{rulik2022control, chung2017performance}, biological signals, like eye gaze and Electromyography EMG~\cite{cio2019proof, fischer2024scoping, cheng2022robotic}, and speech interfaces~\cite{poirier2023efficient}. Traditional speech systems often rely on predefined commands, limiting their ability to handle open-ended, underspecified requests. Recent advances in large language models (LLMs) enable natural language input, and prior work has studied clarification of ambiguous instructions that focus on algorithms~\cite{lin2025ask, gao2022dialfred, ramrakhya2025grounding, jian2025teaching}. However, real-world, task-grounded dialogue datasets for benchmarking complex assistive voice interaction remain limited.
\vspace{-0.5em}

\subsection{Teleoperation}
Teleoperation is widely used for robot data acquisition across platforms, including fixed manipulators~\cite{zhao2023learning, iyer2024open, lin2025learning} and humanoids~\cite{cheng2024open}, but is often designed for local operation due to perception and communication constraints. In assistive robotics, integrated teleoperation of a wheelchair and a WMRA remains rare. Although OpenTeach~\cite{iyer2024open} provides an extensible architecture for manipulators, it is not optimized for coordinated mobile manipulation. 
\vspace{-0.3em}


\section{Integrated Robotic System}
\label{sec:Integrated Robotic System}





\begin{figure}
    \centering
    \includegraphics[width=0.6\linewidth]{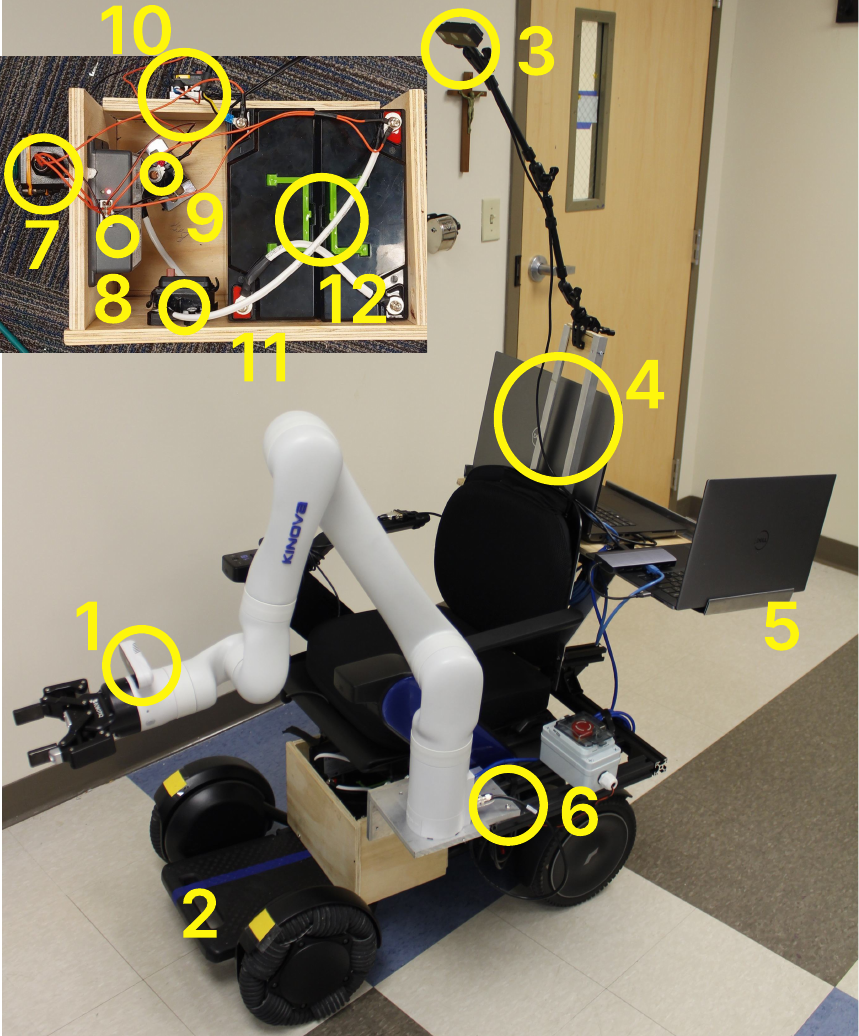}
    \vspace{-0.5em}
    \caption{WheelArm Assistive Platform Hardware Overview: (1) Kinova Gen3 arm (6-DoF) with Intel RealSense D415 camera; (2) WHILL Model CR2 wheelchair; (3) Luxonis OAK-D-W camera and IMU; (4) edge device for teleoperation and data collection; (5) back shelf; (6) robotic arm adapter; (7) precharge switch; (8) precharge module; (9) main contactor; (10, 11) circuit breakers (15~A, 20~A); (12) two 12~V lithium batteries.}
    \label{fig:WheelArm}
\vspace{-0.5em}
\end{figure}

We used a WHILL Model CR2 wheelchair integrated with a 6-DoF Kinova Gen3 robotic arm, referred to as \textbf{WheelArm}, as the testbed for data collection. Details are shown in the \textbf{Fig~\ref{fig:WheelArm}}. The teleoperation and data collection pipelines run on two shelf-mounted laptops: a Dell Precision 5570 (Core i9, 32 GB RAM; Laptop~A) and a Dell Precision 7780 (Core i9, 64 GB RAM; Laptop~B), both running Ubuntu~22.04. The VR teleoperation app runs on a Meta Quest~3S (128 GB).

\section{Teleoperation and Data Collection Framework}
\label{sec:Teleoperation and Data Collection Framework}







\begin{figure}
    \centering    
    \includegraphics[width=0.45\textwidth]{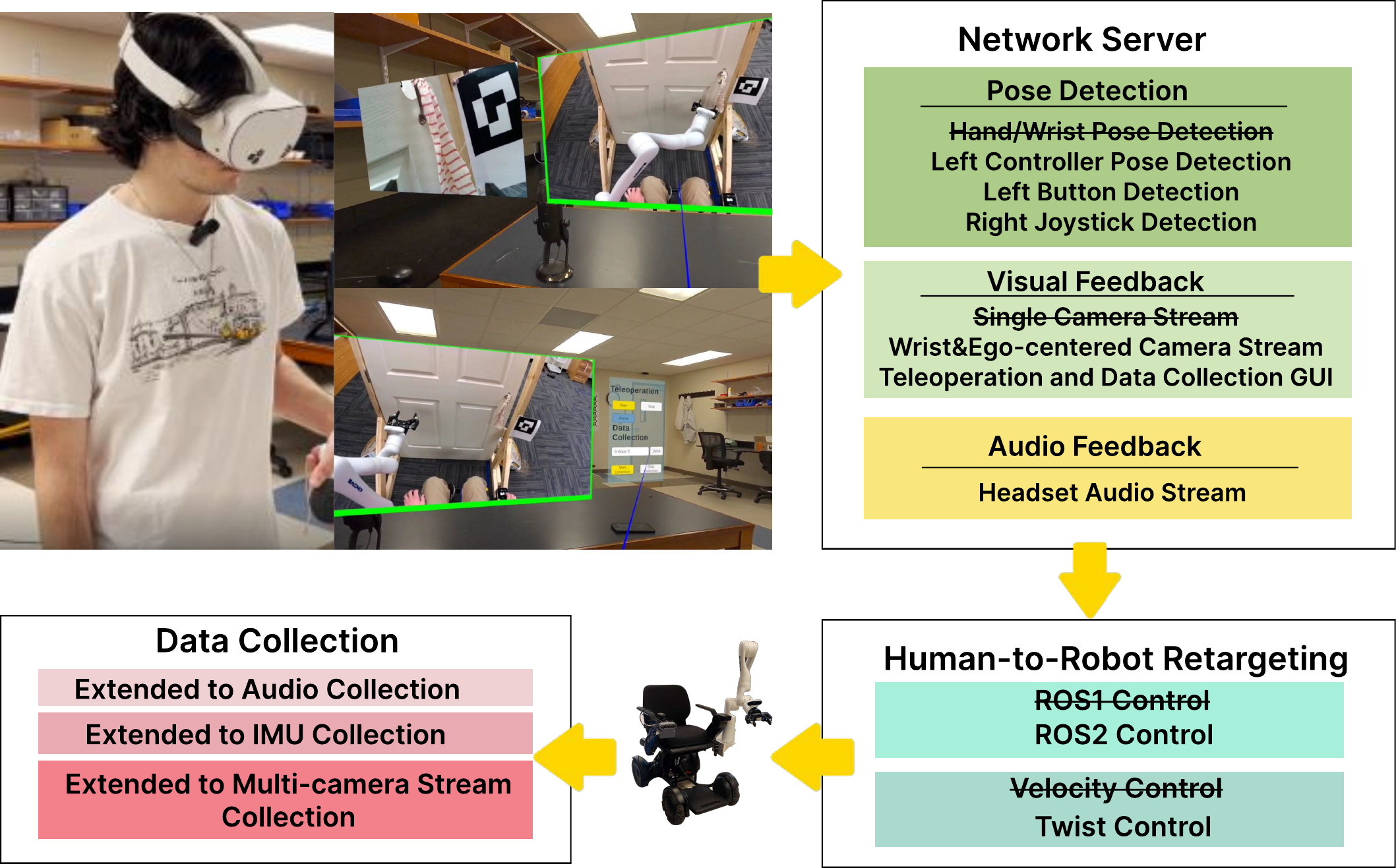}
    \caption{Teleoperation and data collection framework. Strikethrough elements indicate components from the initial software version that were updated or extended in this work.}
    \vspace{-2em}
    \label{fig:tele_frame}
\end{figure}

Built on OpenTeach~\cite{iyer2024open}, an open-source teleoperation system for manipulation, we extended it with a Kinova Gen3 6-DoF arm and Whill Model CR2 wheelchair to develop a teleoperation system. The original OpenTeach~\cite{iyer2024open} is simplified from hand tracking to controller tracking, synchronizing the left-controller motion with the robotic arm movement and the right-hand joystick control with the wheelchair. Additionally, we added button and joystick controls for the gripper and wheelchair. To support the WoZ protocol \cite{riek2012wizard} (see \textbf{Sec. \ref{sec:WoZ Protocol}}), we extended the VR application with a researcher-friendly Graphic User Interface that allows the teleoperator to control the start and stop of data collection without communicating with the on-site researcher. At the sensor level, we replaced the single camera stream with two views: a wrist camera for precise grasping and an ego-centric camera for situational awareness, improving the teleoperator’s view of the environment. To support the integrated execution of the arm and the wheelchair (mobile manipulator), we upgraded the control architecture from Robot Operating System ROS 1 to ROS 2. Arm motion is driven by a twist-based mapping from the left VR controller pose to Cartesian commands, with real-time Inverse Kinematics (IK) solved by MoveIt~2. The wheelchair is commanded via a virtual joystick. The customized teleoperation and multimodal data collection workflow is shown in \textbf{Fig.~\ref{fig:tele_frame}}. Inheriting the data transmission of OpenTeach~\cite{iyer2024open}, which uses TCP/IP and ZeroMQ socket to transfer data from the VR headset to the local laptop for collection, except audio, which is collected by UDP/IP. 

In data collection, we extended from the original single camera stream, robot joint states, and end-effector cartesian pose to two camera streams, audio, and IMU data. The audio recordings are sampled at 48 kHz with 16-bit resolution in mono and saved as \texttt{.wav} files. Camera streams are collected at 15 and 12 fps.

Two laptops are connected via Ethernet. Laptop~A runs the VR teleoperation and data collection pipelines: it receives VR inputs (left-controller pose, gripper buttons, and right-joystick commands), computes end-effector twist commands, and records synchronized streams (video \texttt{.avi}, audio \texttt{.wav}, and robot logs in HDF5). Laptop~B runs the robot-side ROS~2 packages for the wheelchair and the arm. The machines communicate via Cyclone DDS, enabling real-time topic synchronization; robot states are published to Laptop~A for collection while control commands are sent to Laptop~B for execution.

\vspace{-0.8em}

\section{Wizard-of-Oz Protocol}
\label{sec:WoZ Protocol}

WoZ \cite{riek2012wizard} is an experimental paradigm in which subjects interact with a system they believe to be autonomous, while key behaviors are covertly controlled by a human operator. It is commonly used in HRI to elicit realistic user reactions before fully autonomous capabilities are available. In our study, we adopt a WoZ protocol to simulate WheelArm's dialogue-based autonomy and record subjects' natural reactions.

\subsection{WoZ Scheme Setup}
Multiple schemes are set up to implement WoZ. During the experiment, two researchers are involved: one serves as the teleoperator (see \textbf{Fig~\ref{fig:overview}}) in Room B, and the researcher accompanies the subject in Room A. The physical separation conceals the teleoperator, and subjects are told WheelArm is an intelligent robot capable of natural-language, multi-turn dialogue. Communication is routed through a shared Zoom voice channel~\cite{zoom_workplace}, with the teleoperator’s speech converted in real time using the w-okada system~\cite{wokada_voicechanger} to reduce speaker cues and maintain a consistent robot persona.
\vspace{-0.3em}

\section{Experimental Design}
\label{sec:Experimental Design}



\subsection{Experiment Environment}
The experiment is conducted in Rooms A and B. Room A serves as the data collection room, set up with a hand-made door, a drawer, a table for feeding and drinking, and a table for trash and a trash bin. Room B is the teleoperation room, which features a table with a microphone (see \textbf{Fig~\ref{fig:overview}}).

\subsection{Task Design}
Our tasks are selected based on International Classification of Functioning, Disability and Health (ICF) framework \cite{Ustun2003ICF} and online examples of wheelchair users with robotic arms:\\
\vspace{-1em}  
\begin{itemize}
    \item Drinking: Drinking seems simple but requires care near the mouth and attention to user preferences. In Room~A, a can of sparkling water is placed on a table, and subjects instruct WheelArm to bring it to their mouth for self-drinking and then return it.
    

    \item Feeding: Feeding requires selecting, transferring, and safely delivering food near the mouth. We placed multiple differently colored apples and several oranges on a table to simulate a common object-selection scenario.
    
    
    \item Cleaning: Cleaning is similar to pick-and-place but often targets areas outside the robot’s view. We simulated this by having WheelArm throw trash into an out-of-view bin instead of a visible tabletop bin.

    
    \item Drawer Opening: Opening a drawer requires identifying the correct drawer and executing a reliable pull. To support robust two-finger grasping, we attached a cloth strap to the handle, inspired by service-dog assistance techniques~\cite{nguyen2009pps}.
    
    
    \item Door Opening: Door opening is often overlooked in assistive studies, which require coordinated control of both the wheelchair and the arm. To capture this whole-body challenge, we built a wooden door in Room~A and attached a cloth strap to the handle to facilitate grasping with a two-finger gripper.
\end{itemize}
\vspace{-1em}  

\subsection{Participants}
We collected 5 subjects' data whose demographic information can be found in the \textbf{Table~\ref{tab:participants}}. The human subject study protocol followed an approved protocol by Saint Louis University's Institutional Review Board (IRB) (Assurance No. FWA00005304). 


\begin{table}[]
    \centering
    \caption{Subjects Demographic Overview}
    \begin{tabular}{|c|c|c|c|c|}
         \hline
         Total Participants (N) & 5\\
         \hline
         Age (Mean±SD) &  22.8±3.92\\
         \hline
         Gender & 2F/3M\\
         \hline 
         Prior Assistive Technology Experience & None\\
        \hline
    \end{tabular}
    \vspace{-2em}  
    \label{tab:participants}
\end{table}




\vspace{-1em} 

\subsection{Study Procedure}
Before pilot data collection, the teleoperator (a junior reseacher) was trained for over 30 hours (by senior researchers) to be an expert on performing tasks flawlessly. This ensures the teleoperator produces high-quality motion and reduces teleoperation failures. To approximate robot-like partial observability, the teleoperator views the environment only through the wrist and ego-centric camera streams in the VR interface, limiting situational awareness to onboard sensing and prompting robot-plausible clarification and actions.


After screening by the researcher according to an IRB-approved protocol, eligible participants schedule an in-person session in Room~A. Upon arrival, participants review and sign a consent form, HIPAA authorization, and a Demographic and Background Questionnaire. The researcher then introduces WheelArm, described to subjects as an intelligent (autonomous) assistive robot, and explains the five experimental tasks. After data collection is completed, subjects participate in a debriefing in which the WoZ procedure is disclosed, followed by a re-consent step that gives them the right to withdraw their data. After re-consenting, subjects perform a brief manual-control trial to open the door using an Xbox controller and a joystick. Finally, they complete the post-study feedback questionnaire.
\vspace{-0.5em}

\subsection{Questionnaire}
The questionnaire contains three sections. The first Likert-scale table includes five items assessing participants' enjoyment of the dialogue-driven interaction. The second Likert-scale table evaluates perceptions of the system autonomy, using a manual-control condition as a baseline to compare preferences between WheelArm's automated behavior and joystick and Xbox-based control. Finally, the open-ended questions solicit qualitative feedback and suggestions regarding the interaction and autonomy.
\vspace{-0.4em}

\section{Data Processing}
\label{sec:Data Processing}
In this section, we first filter the raw trials to retain only successful teleoperation runs, defined as completing the task without drops or excessive force on non-target objects. We then synchronize the video and whole-body joint streams with the full numerical data using a threshold-based alignment followed by linear interpolation. 

\subsection{Trial Selection}
Failure modes, such as dropping or unintended displacement, are inevitable in teleoperation due to muscle fatigue. To ensure high data quality, we retain only successful trials. A trial is labeled successful only if the task is completed without object drops, items falling, environmental collisions, or inappropriate force. The proportion of filtered tasks relative to the total number of recorded tasks is presented in the \textbf{Table~\ref{tab: Trial Filter}}. 


\begin{table}[]
    \centering
    \caption{Dataset Curation Results: Raw vs. Successful Trials by Task}
    \begin{tabular}{|c|c|c|}
    \hline
         & Num. of Raw Trials & Num. of Successful Trials \\
    \hline
      Cleaning   & 9 & 4 \\
    \hline
    Door Opening &  16 & 15 \\
    \hline
    Drawer Opening & 17 & 16 \\
    \hline
    Drinking & 11 & 9 \\
    \hline
    Feeding & 13 & 9 \\
    \hline   
    Total & 66 & 53 \\
    \hline
    Percentage & -- & 80.30\%\\
    \hline
    \end{tabular}
    \label{tab: Trial Filter}
    \vspace{-2em}  
\end{table}

\subsection{Data Synchronization}
To synchronize the ego-centered (12 fps) and wrist-mounted (15 fps) video streams with the whole-body joint data, we first compute their temporal overlap. The reference start time $t_{\mathrm{ref\_start}}$ and end time $t_{\mathrm{ref\_end}}$ are defined based on the latest of the start times and the earliest of the end times. Since video frames are sampled at discrete intervals, we construct a reference time grid at the lower frame rate (12 fps) over the interval $[t_{\mathrm{ref\_start}},\, t_{\mathrm{ref\_end}}]$ to avoid unnecessary upsampling. For each timestamp on this reference grid, we select the closest frame from each video stream, accepting the match only if the absolute difference between the frame timestamp and the reference time is below a tolerance $\tau$. The matching rule is expressed in the \textbf{Equation~\ref{Equ: Matching Rule}}.


\begin{equation}
\hat{i}_k^{(c)} \;=\;
\begin{cases}
i_k^{(c)}, & \text{if } \left| t^{(c)}_{i_k^{(c)}} - t_k \right| \le \tau,\\[2pt]
\hat{i}_{k-1}^{(c)}, & \text{otherwise,}
\end{cases}
\label{Equ: Matching Rule}
\end{equation}
where $c$ denotes the camera stream, $k$ indexes the reference time step, $t$ indexes timestamps and $i \in \{1,\dots,N_c\}$ indexes the original frames of stream $c$. The index $i_k^{(c)}$ is the nearest-neighbor frame to $t_k$, and $\hat{i}_k^{(c)}$ is the final selected frame index after applying the tolerance rule (with $\hat{i}_{k-1}^{(c)}$ the previously selected index).

Numerical data, including end-effector Cartesian states, IMU data, robotic arm joint states, wheelchair wheel speeds, and positions, are also interpolated with the reference time for future denoising.
\vspace{-0.5em} 

\subsection{Data Denoising}
To reduce sensor noise, we applied a zero-phase 4th-order Butterworth low-pass filter to the end-effector Cartesian pose and whole-body joints (arm joints and wheelchair wheels) with a 5 Hz cutoff, preserving teleoperator motion while attenuating high-frequency artifacts. We applied the same filter to the IMU with a 10 Hz cutoff.
\vspace{-0.5em} 

\subsection{Linguistic Processing}
The raw audio streams were first transcribed using OpenAI's Whisper model~\cite{radford2023robust} to generate initial text logs. Given the noisy real-world environment, such as motor noise from the wheelchair, the automated transcripts are manually verified and corrected by human annotators to match the subject's audio, keeping the grammar mistakes, verbal slips, incomplete sentences, and not well-phrased expressions for realism. Each utterance is then sequenced into multi-turn conversations for future analysis.
\vspace{-0.6em}

\section{Data Analysis and Results}
\label{sec:Data Analysis}





The results here will be answering the three research questions below:
\begin{itemize}
    \item Q1: Can our data collection framework using WoZ effectively collect expert robot data for the assistive tasks?
    \item Q2: What are the characteristics of dialogues in assistive contexts, and to what extent does clarification dialogue resolve ambiguities inherent in these utterances?
    \item Q3: What is the participants' feedback on this dialogue-based interaction?
\end{itemize}
\vspace{-0.6em} 

\subsection{Data Quality (Q1)}
\subsubsection{Analysis}
To evaluate whether our data collection framework gathers high-quality expert data, we analyzed task performance with task time and end-effector path length (mean and standard deviation SD), and robot movements' smoothness by jerk analysis (mean jerk of the end-effector and the wheelchair). This analysis assesses the robot's data quality in terms of smoothness and human comfort.

Linear Jerk, defined as the third derivative of the positions $p$ and the rate at which acceleration changes. Its magnitude reflects the robot's motion fluidity. We compute the mean jerk for the end-effector ($J_{\mathrm{mean}}^{ee}$) and the wheelchair ($J_{\mathrm{mean}}^{w}$) on both the trial level and the task level. For each trial, given Cartesian positions $\mathbf{p}(t_k)$ sampled at timestamps $\{t_k\}_{k=1}^{N}$, the jerk magnitude at sample $k$ is defined as $j_k=\lVert \dddot{\mathbf{p}}(t_k)\rVert_2$
where $k$ indexes time samples within the trial, $\mathbf{p}$ is the Cartesian position (EE position for $J_{\mathrm{mean}}^{ee}$ and wheelchair position for $J_{\mathrm{mean}}^{w}$), and $\lVert\cdot\rVert_2$ denotes the Euclidean norm. The trial-level mean jerk is computed by averaging $j_k$ over the $N$ samples:
\vspace{-0.5em}
$$
J_{\mathrm{mean}}^{(i)}=\frac{1}{N}\sum_{k=1}^{N} j_k,
$$
where $i$ indexes trials.

Given the set of trial-level mean jerks $\{J_{\mathrm{mean}}^{(i)}\}_{i=1}^{n_{\mathrm{trial}}}$ for a task with $n_{\mathrm{trial}}$ trials, the task-level mean is
$$
J_{\mathrm{mean}}^{\mathrm{task}} = \frac{1}{n_{\mathrm{trial}}}\sum_{i=1}^{n_{\mathrm{trial}}} J_{\mathrm{mean}}^{(i)},
$$
and the task-level (sample) standard deviation is
$$
J_{\mathrm{SD}}= \sqrt{\frac{1}{n_{\mathrm{trial}}-1}\sum_{i=1}^{n_{\mathrm{trial}}}\left(J_{\mathrm{mean}}^{(i)}-J_{\mathrm{mean}}^{\mathrm{task}}\right)^2}.
$$
The task-level jerk is expressed as $J_{\mathrm{mean}}^{\mathrm{task}} \pm J_{\mathrm{SD}}$.

For the qualitative analysis, we present time-series plots from two cameras to illustrate visual data and relative dialogues from one of the experiments for each task.



\subsubsection{Results}

\begin{figure*}
    \centering    
    \includegraphics[width=\textwidth]{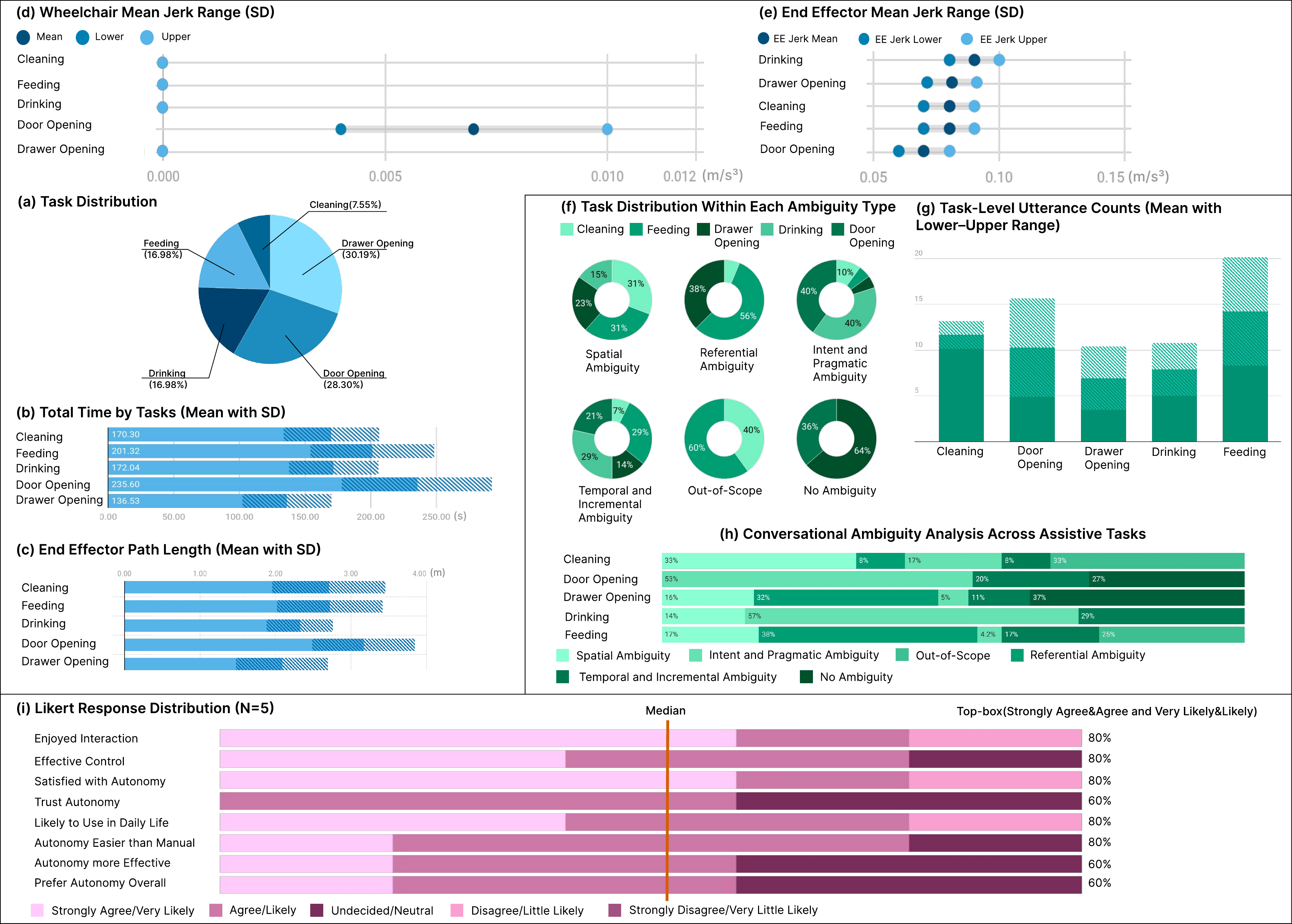}
    \vspace{-1em}
    \caption{Quantitative Analysis: (a-e) Task Performance Analysis: (a) is the task distribution in the pilot dataset, (b) is total completion time (mean $\pm$ SD) by task, (c) is end-effector (EE) path length in meters (mean $\pm$ SD) across the five tasks, capturing both average motion extent and variability, (d) is wheelchair mean jerk (mean $\pm$ SD); nonzero jerk is primarily observed in the door-opening task, where the wheelchair is repositioned, while remaining smooth, and (e) is EE mean jerk (mean $\pm$ SD), indicating consistently low jerk and thus smooth, unjittered EE motion during data collection. (f-h) Dialogue Analysis: (f) is task distribution in each ambiguity, (h) is the dialogue-ambiguity distribution for each task, and (g) is utterance counts of each task. User Feedback Analysis: (i) Likert Response Distribution demonstrates how subjects like this dialogue-based interaction and an intelligent robot on assistive tasks.}
    \label{fig: quan analysis}
\vspace{-0.5em}
\end{figure*}

\textbf{Fig.~\ref{fig: quan analysis}, parts a-e} (all blue plots) answer Q1 by showcasing task distribution (\textbf{Fig.~\ref{fig: quan analysis} - a}), time by task (\textbf{Fig.~\ref{fig: quan analysis} - b}), end-effector path length (\textbf{Fig.~\ref{fig: quan analysis} - c}), and jerk analysis (\textbf{Fig.~\ref{fig: quan analysis} - d, e}) to prove the smooth, kinematically consistent, and reliable movement of the collected data. 

Our pilot dataset comprises 53 trials in total with cleaning (n=4), door opening (n=15), drawer opening (n=16), drinking (n=9), and feeding (n=9) (\textbf{Fig.~\ref{fig: quan analysis}-a}). Quantitative analysis (\textbf{Fig.~\ref{fig: quan analysis}-d, e}) confirms high motion smoothness and low vibration across all tasks. Wheelchair jerk magnitudes remained well below the established 0.3--0.9 $m/s^3$ ~\cite{bae2019toward} comfort baseline for public transportation, indicating exceptionally fluid movement. While no direct standard exists for robotic arm jerk, the ISO 2631-1:1997 standard~\cite{mccallig2021whole} defines the threshold for human whole-body comfort at $0.315 m/s^2$. Vibration magnitudes below this value are categorized as comfortable. As shown in \textbf{Fig. \ref{fig: quan analysis} - e}, the peak mean ($\approx0.1 m/s^3$) implies reaching the ISO discomfort threshold ($0.315 m/s^2$) would require over 3 seconds of continuous, unidirectional acceleration. Since assistive tasks consist of short, multi-directional maneuvers, the system effectively maintains comfortable acceleration levels and avoids the jerky impulses.

\begin{figure*}
    \centering
    \includegraphics[width=0.95\textwidth]{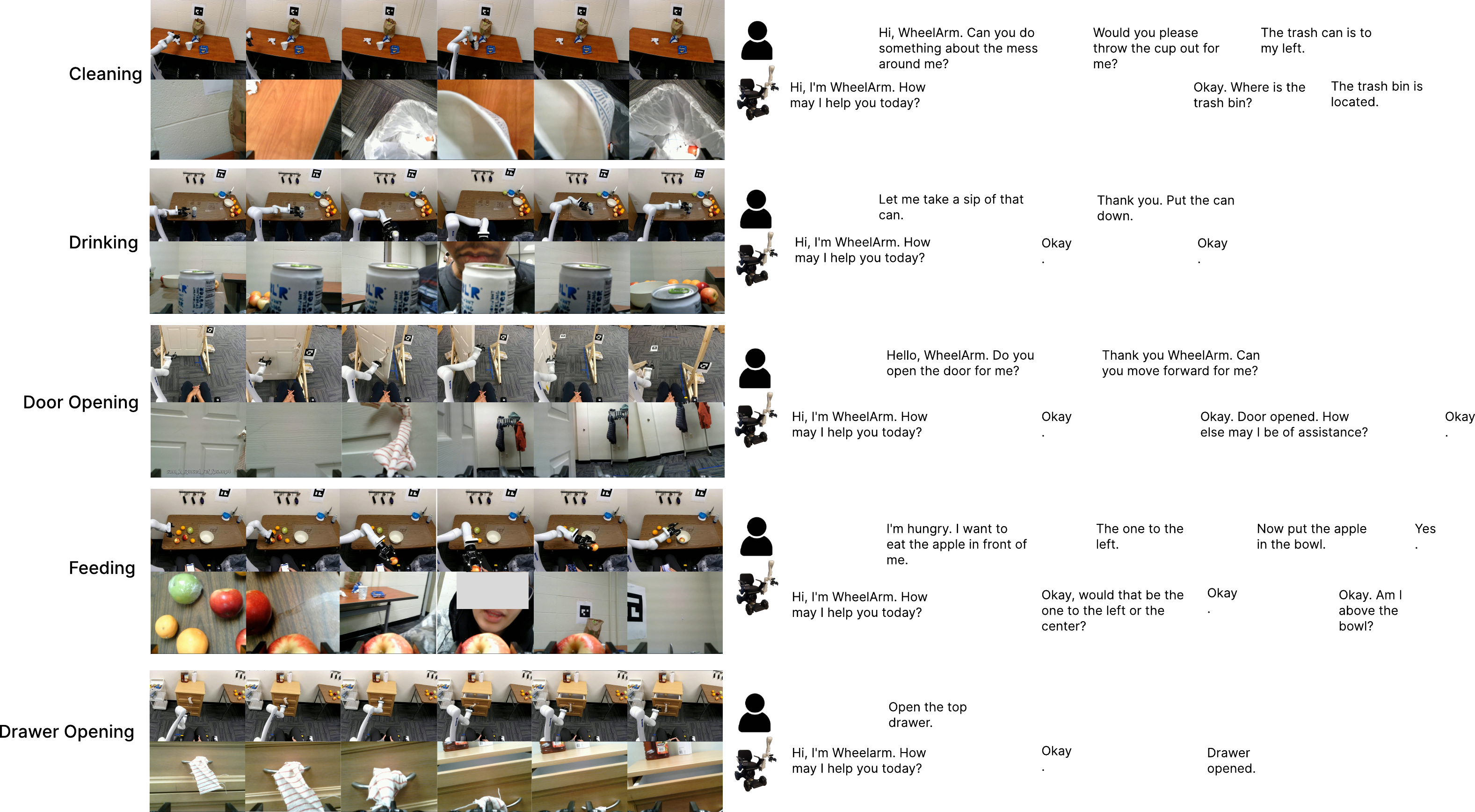}
    \caption{Task Performance Qualitative Demonstration.}
    \label{fig: task qua analysis}
    \vspace{-0.8em}  
\end{figure*}

The qualitative analysis presents videos and conversations from a sample of each task to assess video quality, WheelArm motions, and dialogue between the participants and the WheelArm. \textbf{Fig.~\ref{fig: task qua analysis}} demonstrates high-quality multimodal data collected by our framework. 
\vspace{-0.6em} 

\subsection{Dialogue Ambiguity (Q2)}
\subsubsection{Analysis}
To characterize collected dialogues, we annotate ambiguity types in each trial using the following classification:

\begin{itemize}
    \item Command Clarity: User commands were classified as either \textit{Specific} or \textit{Ambiguous}. A \textit{Specific} command provides sufficient task-level information for autonomous execution by specifying the intended action and an identifiable target in the current context, while leaving low-level motion parameters such as grasp pose to the WheelArm.

    \item Ambiguity Type: For ambiguous commands, we further classified the source of uncertainty into four categories:
    \begin{enumerate}
    \item Spatial Ambiguity~\cite{liu2010ambiguitie}: Uncertainty arising from deictic references. This includes perspective-dependent directions, such as user or robot. For example, the command \textit{Place it far away from me} doesn't provide specific directions.
    \item Referential Ambiguity~\cite{matuszek2014learning}: The inability to uniquely identify a target in cluttered scenes. 
    \item Intent and Pragmatic Ambiguity~\cite{matuszek2014learning}: Vague high-level goals such as \textit{I'm hungry} or indirect speech acts like \textit{I'm thirsty} that lack explicit execution details. 
    \item Temporal and Incremental Ambiguity: Uncertainty regarding the timing or repetition of actions such as again or more. 
    \item Out-of-Scope: When the target is outside the robot’s field of view, additional dialogue is often required to ground references and guide motion under sensing limitations.
\end{enumerate}
\end{itemize}
This annotation process correlates specific linguistic ambiguity with the multimodal context. The transcribed multi-turn dialogues, corresponding image frames, and ambiguity annotations are stored in a \texttt{.jsonl} file for analysis.

\subsubsection{Results}
\textbf{Fig.~\ref{fig: quan analysis}, parts f-h} summarizes the dialogue analysis across tasks. \textbf{Fig.~\ref{fig: quan analysis} - h} compares the dominant ambiguity types by task: cleaning is primarily characterized by spatial and out-of-scope ambiguity (approximately equally frequent); door opening and drinking are dominated by intent and pragmatic ambiguity; drawer opening contains the largest proportion of unambiguous utterances; and feeding exhibits the most referential ambiguity. \textbf{Fig.~\ref{fig: quan analysis} - f} shows the task distribution for each ambiguity type, and \textbf{Fig.~\ref{fig: quan analysis} - g} reports utterance counts at the task level. Although the number of dialogue turns varies across participants, drawer opening generally requires the fewest utterances to specify the command, whereas feeding requires the most to convey the intent clearly to the WheelArm. 
\vspace{-0.7em}

\subsection{Participant Feedback (Q3)}
\subsubsection{Analysis}
All Likert scales range from strongly agree or very likely to strongly disagree or very unlikely, scored from 5 to 1, respectively. We reported the median to summarize participants’ attitudes for each question. We also computed the top-box percentage (ratings of 4 or 5) to quantify the proportion of the most positive responses.
\subsubsection{Results}
Questionnaire results indicate that participants generally enjoyed the dialogue-driven interaction and perceived the system autonomy positively. From \textbf{Fig.~\ref{fig: quan analysis} - i}, the medians reflect strong agreement for enjoyment of the interaction and satisfaction with the autonomy, with other items tending toward agree or likely. Top-box results further support this trend with 5 assessment questions achieving 80\% top-box (ratings of 4-5) and 3 achieving 60\%.
In the open-ended responses, most participants reported enjoying using WheelArm, while one participant suggested that the robot need to move faster. 
\vspace{-0.6em}


\section{Conclusion and Future Work}
\label{sec:Discussion and Future Work}\vspace{-0.2em}
In this work, we presented a multimodal data collection framework for mobile manipulators that enables dialogue-driven interaction to clarify utterance ambiguity. The framework simulates an intelligent robot protocol, combining a VR-based teleoperation and data collection pipeline for real-time collection. The experiments adopt a two-room WoZ protocol to elicit natural multi-turn conversations while maintaining safe and reliable robot execution. Our work addresses the limitations of current \textit{instruction-following} research, which largely relies on single-turn commands and overlooks the underspecification and ambiguity inherent in natural human conversation. By capturing natural conversational nuances grounded in a mobile manipulator, this framework enables the development of intelligent agents for assistive robotics, facilitating a shift toward intuitive assistive devices that go beyond simple execution to resolve intent through dialogue, thereby restoring functional independence in complex, unstructured environments.

In future work, we will scale the study to a larger, more diverse participant pool, expand the tasks to include complex assistive tasks, include haptic feedback, and develop a conversational vision–language–action model that jointly reasons over dialogue, visual context, and robot state to enable ambiguity-aware autonomous assistive behavior.

\vspace{-0.8em}

\maketitle
\thispagestyle{empty}
\pagestyle{empty}



\addtolength{\textheight}{-12cm}   








\end{document}